\documentclass[runningheads]{llncs}
\usepackage{graphicx}
\usepackage{amsmath,amssymb} 
\usepackage{color}
\usepackage{makecell}

\begin{document}
\title{Multi-Domain Pose Network for Multi-Person Pose Estimation and Tracking} 

\titlerunning{Multi-Domain Pose Network}
%
\author{Hengkai Guo\inst{1}\thanks{Corresponding author, email: \email{guohengkai@bytedance.com}} \and
Tang Tang\inst{1} \and
Guozhong Luo\inst{1} \and
Riwei Chen\inst{1} \and
Yongchen Lu\inst{1} \and
Linfu Wen\inst{1}
}
%
\authorrunning{Guo et al}
%

\institute{ByteDance AI Lab}
\maketitle              
\begin{abstract}
Multi-person human pose estimation and tracking in the wild is important and challenging. For training a powerful model, large-scale training data are crucial. While there are several datasets for human pose estimation, the best practice for training on multi-dataset has not been investigated. In this paper, we present a simple network called Multi-Domain Pose Network (MDPN) to address this problem. By treating the task as multi-domain learning, our methods can learn a better representation for pose prediction. Together with prediction heads fine-tuning and multi-branch combination, it shows significant improvement over baselines and achieves the best performance on PoseTrack ECCV 2018 Challenge without additional datasets other than MPII and COCO.
\end{abstract}
\section{Introduction}
Multi-person human pose estimation is an important component in many applications, such as video surveillance and sports video analytics. Though great progress has been made in this field \cite{Cao2017Realtime}\cite{Chen2017Cascaded} thanks to the development of convolutional neural networks (CNNs), human pose estimation remains a challenging problem due to complex poses, diverse appearance, different scales, severe occlusion and crowds. For tracking in videos, the strong camera motions and extreme proximity of people \cite{Andriluka2017PoseTrack} make it even more difficult. 

Similar to other computer vision tasks dominated by deep learning, large-scale training data are crucial to exploit the representation power of CNNs for human pose estimation. There exists several extensive datasets such as COCO Dataset\cite{Lin2014Microsoft}, MPII Dataset\cite{Andriluka20142D}, and PoseTrack Dataset\cite{Andriluka2017PoseTrack}. These datasets differ from each other about the distributions of images, poses and annotation standards. To promote the performance of models, many methods \cite{Girdhar2018Detect}\cite{Xiao2018Simple}\cite{Jin2017Towards} choose to utilize multiple datasets for training. Most of them trained the models on COCO dataset first and then fine-tuned them on PoseTrack dataset\cite{Xiao2018Simple} or MPII dataset\cite{Jin2017Towards}. However, it is still unclear what is the best practice to learn a model from multiple datasets for human pose estimation.

In this paper, we treat the task of training on multi-datasets as multi-domain learning \cite{xiao2016learning}\cite{nam2016learning} and propose a CNN architecture named Multi-Domain Pose Network (MDPN). The network has a common backbone to share the representation from multiple domains and separate prediction heads for dataset-specific pose estimation. During training, we first jointly optimize on all datasets to learn the generic pose embedding. Then each heads are fine-tuned on each domains to further improve the accuracy of localization. We also investigate the prediction strategies for better performance. Evaluated on PoseTrack dataset, our methods with simple network structures significantly improve the learning on multi-dataset over baseline. Moreover, our methods is runner-up of the PoseTrack ECCV 2018 challenge of pose estimation but achieved the best performance without using extra training data other than MPII and COCO datasets. 

\begin{figure}[htb]
\begin{minipage}[b]{0.55\textwidth}
\centering
{\includegraphics[width=\textwidth]{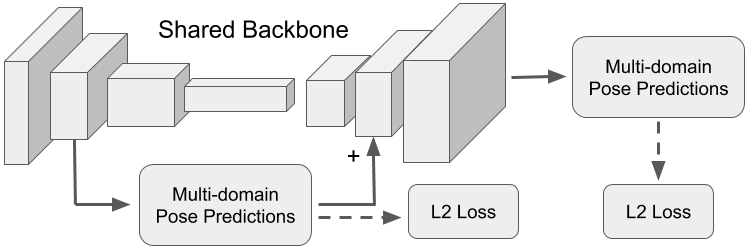}}
\end{minipage}
\begin{minipage}[b]{0.38\textwidth}
\centering
{\includegraphics[width=0.7\textwidth]{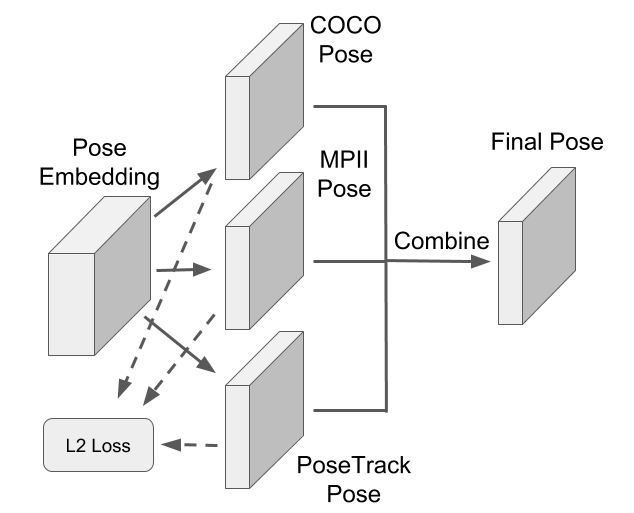}}
\end{minipage}
\label{fig_overview}
\caption{Network overview (Left) and multi-domain prediction (Right).}
\end{figure}

\section{Methods}
Overall, we adopt the top-down approach \cite{Chen2017Cascaded}\cite{Xiao2018Simple} to human pose estimation using only single frame information, which employs person detector to detect all the people in the image and then use single person pose estimator (SPPE) to obtain the human poses for all the boxes. For SPPE, we take advantage of multiple dataset information to train a multi-domain network. After that, we use simple matching \cite{Xiao2018Simple} on adjacent frames to associate individuals into track-lets. 

\subsection{Multi-Domain Pose Network (MDPN)}
For training on multiple datasets, there are three simple solutions:

\textbf{Mixed. } All the datasets are merged into one single dataset. We also merge all the joint sets into one single joint set with a total number of 21 keypoints. During training, gradients are only back-propagated to the annotated joints for each sample. Mixing the datasets can make full use of all the information from all datasets, but different annotation standards for datasets on the same joint may distract the training procedure.

\textbf{Transfer learning. } As done in \cite{Jin2017Towards}\cite{Xiao2018Simple}, we can also train the model on one dataset first to learn the generic representation. Then fine-tuning is performed on the target dataset. COCO dataset\cite{Lin2014Microsoft} is often selected to pre-train the models because its data distribution is good to train a generic pose estimation model. Transfer learning can speed up the learning and often achieves good results. But the learnt embedding is suboptimal for pose estimation and it is easy to lose the knowledge from the first dataset when training on the target dataset for a long time as observed by \cite{zhu2017multi}. 

\textbf{Multi-domain learning. } Another approach to train on multi-dataset is to view it as a multi-domain learning task like \cite{nam2016learning}. It uses a common backbone network to learn a common pose representation and several prediction heads to learn domain-specific pose estimation. Compared with mixed datasets, multi-domain learning addresses the different annotation distribution problem. But to balance between different domains, the learnt prediction heads may not be optimal. Moreover, the predictions of multi-domain network only use one single head, which is a waste of information from other datasets.

According to the analysis above, we propose Multi-Domain Pose Network to solve such problems. We first apply multi-domain learning on all datasets. Then we fine-tune the full model on COCO dataset to optimize the embedding and COCO head. Finally we fix the backbone together with COCO head and fine-tune our network on the combination of MPII and PoseTrack dataset for the remain heads. Fig. \ref{fig_overview} illustrates the whole network structure. The details of structure will be explained in Section. \ref{sec_impl}.

For prediction, one simple strategy is to use the predictions from corresponding dataset head. In order to exploit all the information from different datasets, we combine the predictions from different heads to form the final estimation (Fig. \ref{fig_overview}). There are several ways of combination, which will be discussed and compared in Section \ref{sec_expr}. Such methods can also be viewed as a lightweight multi-dataset ensemble implemented by multi-branch predictions like \cite{Li2016Convolutional}\cite{Guo2017Region}.

\subsection{Implementations}
\label{sec_impl}
\textbf{Model structure. } We use ResNet-152 \cite{He2016Deep} with three deconvolution layers as backbone \cite{Xiao2018Simple}. To address vanishing gradients, we add an intermediate prediction after conv3 layers for supervision, and add it back to the second deconvolution layer as skip connection. The size of input image is 384$\times$288.

\textbf{Training. } The cropping and augmentations are the same as \cite{Chen2017Cascaded}. The Gaussian maps with sigma 9 are used as targets. We use the pre-trained models on ImageNet for ResNet backbone. The base learning rate is 0.001 with batch size 128 and Adam optimizer. For the jointly training stage of MDPN, we use 120 epochs. The learning rate is dropped to 0.0001 at 90 epochs. Then we perform 15 epochs for fine-tuning on COCO dataset. Finally we fine-tune the model on MPII and PoseTrack datasets for 20 epochs (The learning rate is dropped to 0.00001 at 10 epochs). To improve the performance of hard keypoints, we change the L2 loss to Online Hard Keypoints Mining (OHKM)\cite{Chen2017Cascaded} loss with 8 top keypoints at 100 epochs. For other models, we follow the training scheme in \cite{Xiao2018Simple}.

\textbf{Testing. } We follow the common practice in \cite{Chen2017Cascaded} with flipping testing and quarter offsets. We also re-score the box with the production of box score and average keypoint scores \cite{Chen2017Cascaded} after predictions.

\textbf{Detection. } We use four public person detectors trained on COCO dataset \cite{Lin2014Microsoft}, including Faster R-CNN \cite{Ren2015Faster}, Mask R-CNN \cite{He2017Mask}, YOLO \cite{Redmon2018YOLOv3}, and DCN \cite{Dai2017Deformable}. Then we merge all the boxes with NMS of 0.6 and use them as detection results.

\textbf{Tracking. } We follow the pipelines of flow-based tracking \cite{Xiao2018Simple} with four modifications. First, we apply OKS-NMS \cite{Papandreou2017Towards} of 0.4 after pose estimation. Second, we use Hungarian matching instead of greedy matching. Third, after tracking we prune short track-lets that contain less than 2 frames to reduce the false positive cases. Finally, we do not employ box propagation because the detector ensemble is strong enough. For multi-frame flow tracking, we use at most 8 frames before.

\section{Experiments}
\subsection{Datasets and Evaluation}
We train our models on three datasets: COCO-2017 dataset \cite{Lin2014Microsoft}, MPII Dataset \cite{Andriluka20142D}, and PoseTrack-2018 Dataset \cite{Andriluka2017PoseTrack}. Then we evaluate our methods on the PoseTrack-2018 validation dataset. For multi-person pose estimation, we use mean Average Precision (mAP) metric. For multi-person tracking, we use Multiple Object Tracking Accuracy (MOTA) metric. To compare with state-of-the-art methods, we also evaluate our methods on PoseTrack-2017 validation dataset. For ablation study, we construct a min-val dataset from PoseTrack-2018 validation dataset by uniformly sub-sampling 15 sequences out of 75 sequences. 

\subsection{Ablation Study}
\label{sec_expr}
ResNet-50 of input 256$\times$192 without skip connection is used here for simplicity.

\begin{table*}[htb]\tiny
\caption{Different training (Left) and testing (Right: MDPN-B without fine-tuning) methods on PoseTrack-2018 \textbf{min-val} dataset with ResNet-50.}
\label{table_res50_posetrack}
\begin{minipage}[b]{0.48\textwidth}
\centering
\begin{tabular}{c|c|c|c}
\hline
Methods & \makecell{Wrist \\ mAP} & \makecell{Ankle \\ mAP} & \makecell{Total \\ mAP} \\
\hline
MPII & 58.3 & 49.1 & 58.7 \\
COCO & 70.8 & \textbf{59.2} & 68.4 \\
PoseTrack & 53.9 & 43.2 & 55.5 \\
COCO $\to$ PoseTrack & 67.5 & 56.4 & 66.8 \\
COCO $\to$ PoseTrack + MPII & 68.5 & 57.1 & 68.0 \\
Mixed & 67.1 & 52.5 & 66.3 \\
MDPN-B w/o FT & 68.4 & 53.7 & 67.7 \\
MDPN-B & \textbf{71.8} & 56.9 & \textbf{70.7} \\
\hline
\end{tabular}
\end{minipage}
\begin{minipage}[b]{0.48\textwidth}
\centering
\begin{tabular}{c|c|c|c}
\hline
Methods & \makecell{Wrist \\ mAP} & \makecell{Ankle \\ mAP} & \makecell{Total \\ mAP} \\
\hline
COCO branch             & 68.8 & 54.6 & 66.0 \\
PoseTrack branch        & 68.0 & 52.9 & 66.7 \\
COCO + PoseTrack branch & \textbf{69.1} & 54.6 & 67.4 \\
COCO + MPII branch (A)  & 69.0 & \textbf{54.7} & \textbf{67.7} \\
Voting (B)              & 68.4 & 53.7 & \textbf{67.7} \\
\hline
\end{tabular}
\end{minipage}
\end{table*}

\textbf{Testing. } We have tried different combination methods on the multi-domain model: 1) \emph{COCO branch}: Using the COCO branch and interpolating the head positions from other keypoints. 2) \emph{PoseTrack branch}: Using the PoseTrack branch. 3) \emph{COCO + PoseTrack branch}: Using the COCO branch with the head position from PoseTrack branch. 4) \emph{COCO + MPII branch}: Using the COCO branch with the head position from MPII branch. 5) \emph{Voting}: Averaging the heatmaps from common keypoints from all branches. From the right part of Table. \ref{table_res50_posetrack}, the last two methods achieve the best performance. So we will only use these two methods for remain testing and refer them as method A and B. 

\textbf{Training. } We compare different training strategies for multi-dataset: 
1) \emph{MPII}: Training on MPII. 2) \emph{COCO}: Training on COCO. 3) \emph{PoseTrack}: Training on PoseTrack. 4) \emph{COCO$\to$ PoseTrack}: Training on COCO and fine-tuning on PoseTrack \cite{Girdhar2018Detect}\cite{Xiao2018Simple}. 5) \emph{COCO$\to$ PoseTrack + MPII}: Training on COCO and fine-tuning on mixed datasets of MPII and PoseTrack \cite{Jin2017Towards}. 6) \emph{Mixed}: Training on mixed dataset. 7) \emph{MDPN-B w/o FT}: Training with multi-domain learning without fine-tuning and testing with method B. 8) \emph{MDPN-B}: Training with multi-domain learning with fine-tuning and testing with method B. 

Left part of Table. \ref{table_res50_posetrack} shows that among all approaches, the MDPN-B achieves the best performance. And fine-tuning after multi-domain training is important for the final performance (+3.0 mAP). As for the results of single dataset, training on COCO performs the best even without head annotations, while the accuracy of PoseTrack is worst. This is because the images in PoseTrack are obtained from limited videos and contain duplicate information, which leads to a smaller dataset. Another conclusion is that fine-tuning does not always improve the performance on the target dataset due to the knowledge forgetting problems.

\textbf{Post-processing. } Table. \ref{table_post_posetrack} indicates that all post-processing is necessary for final performance. The OKS-NMS is crucial for mAP because too many false positive part detections may mislead the matching stage of evaluation. 

\begin{table}[htb]\tiny
\caption{Different post-processing methods for pose estimation (Left) and tracking (Right) on PoseTrack-2018 \textbf{min-val} dataset.}
\label{table_post_posetrack}
\begin{minipage}[b]{0.4\textwidth}
\centering
\begin{tabular}{c|c}
\hline
Methods & Total mAP \\
\hline
MDPN-A & \textbf{70.7} \\
w/o Gaussian filter & 70.3 \\
w/o quarter offset & 70.5 \\
w/o box threshold & 70.0 \\
w/o OKS-NMS & 51.2 \\
w/o box re-score & 70.2 \\
\hline
\end{tabular}
\end{minipage}
\begin{minipage}[b]{0.6\textwidth}
\centering
\begin{tabular}{c|c|c}
\hline
Methods & Total mAP & Total MOTA \\
\hline
MDPN-A$^*$ &  61.5 & \textbf{52.6} \\
w/o box threshold & 61.5 & 46.0 \\
w/o keypoint threshold & \textbf{67.6} & 28.3 \\
w/o track-let pruning & 61.7 & 52.3 \\
w/o flow track & 61.4 & 52.2 \\
\hline
\end{tabular}
\end{minipage}
\end{table}

\subsection{Results on PoseTrack datasets}
We evaluate our methods on PoseTrack 2017 \cite{Girdhar2018Detect}\cite{xiu2018pose}\cite{Xiao2018Simple} and 2018 dataset \cite{fang2017rmpe}. We use AlphaPose \cite{fang2017rmpe} model as baseline and apply branch combination, OKS-NMS and re-scoring (\emph{AlphaPose++}). 

 Table. \ref{table_result_posetrack} and Table. \ref{table_result_posetrack_track} show all the results on validation sets. For 2017 dataset, our methods show comparable performance with state-of-the-art method\cite{Xiao2018Simple} and outperform other methods. For 2018 dataset, our methods also surpass the baselines with large margin. Meanwhile, testing with COCO-MPII combination is better than that with voting for ResNet-152. 

  For test set (Table. \ref{table_result_challenge}), our MDPN methods beats all the other methods trained only on COCO, MPII and PoseTrack by a large margin (7.0 mAP for no-tracking and 3.2 mAP for tracking). The no-tracking version also achieves the second best performance among all methods. For tracking, our method also performs the best among all methods without extra datasets for MOTA by a large margin (3.6 MOTA) and achieves the third best accuracy in all methods.

\begin{table*}[htb]\tiny
\caption{mAP on PoseTrack 2017 and 2018 datasets. * means with tracking.}
\label{table_result_posetrack}
\centering
\begin{tabular}{c|c|c|c|c|c|c|c|c|c}
\hline
Methods & Dataset  & \makecell{Head \\ mAP} & \makecell{Shoulder \\ mAP} & \makecell{Elbow \\ mAP} & \makecell{Wrist \\ mAP} & \makecell{Hip \\ mAP} & \makecell{Knee \\ mAP} & \makecell{Ankle \\ mAP} & \makecell{Total \\ mAP} \\
\hline
Detect-and-Track\cite{Girdhar2018Detect} & val17 & 67.5 & 70.2 & 62.0 & 51.7 & 60.7 & 58.7 & 49.8 & 60.6 \\
PoseFlow\cite{xiu2018pose} & val17 & 66.7 & 73.3 & 68.3 & 61.1 & 67.5 & 67.0 & 61.3 & 66.5 \\
ResNet-152\cite{Xiao2018Simple} & val17 & 81.7 & 83.4 & 80.0 & 72.4 & 75.3 & 74.8 & 67.1 & 76.7 \\
MDPN-152-A & val17 & 85.2 & 88.5 & 83.9 & 77.5 & 79.0 & 77.0 & 71.4 & \textbf{80.7} \\
MDPN-152-A$^*$ & val17 & 79.8 & 84.8 & 78.6 & 71.7 & 74.8 & 72.3 & 67.5 & 75.8 \\
\hline
AlphaPose\cite{fang2017rmpe} & val18 & 63.9 & 78.7 & 77.4 & 71.0 & 73.7 & 73.0 & 69.7 & 71.9 \\
AlphaPose++   & val18 & 71.9 & 79.7 & 78.3 & 71.7 & 74.3 & 73.3 & 70.1 & 74.0 \\
MDPN-50-B     & val18 & 72.6 & 75.7 & 75.8 & 69.7 & 72.1 & 70.2 & 65.4 & 71.7 \\
MDPN-152-B    & val18 & 76.6 & 77.9 & 76.0 & 69.8 & 68.6 & 70.9 & 67.0 & 72.7 \\
MDPN-152-A    & val18 & 75.4 & 81.2 & 79.0 & 74.1 & 72.4 & 73.0 & 69.9 & \textbf{75.0} \\
MDPN-152-A$^*$  & val18 & 72.4 & 79.0 & 75.3 & 69.6 & 69.2 & 69.2 & 66.7 & 71.7 \\
\hline
\end{tabular}
\end{table*}

\begin{table*}[htb]\tiny
\caption{MOTA on PoseTrack 2017 and 2018 datasets. * means with tracking.}
\label{table_result_posetrack_track}
\centering
\begin{tabular}{c|c|c|c|c|c|c|c|c|c|c|c|c}
\hline
Methods & Dataset & \makecell{Head \\ MOTA} & \makecell{Shou \\ MOTA} & \makecell{Elbow \\ MOTA} & \makecell{Wrist \\ MOTA} & \makecell{Hip \\ MOTA} & \makecell{Knee \\ MOTA} & \makecell{Ankle \\ MOTA} & \makecell{Total \\ MOTA} & \makecell{Total \\ MOTP} & \makecell{Total \\ Prec} & \makecell{Total \\ Rec} \\
\hline
D\&T\cite{Girdhar2018Detect} & val17 & 61.7 & 65.5 & 57.3 & 45.7 & 54.3 & 53.1 & 45.7 & 55.2 & 61.5 & 66.4 & 88.1 \\
PoseFlow\cite{xiu2018pose} & val17 & 59.8 & 67.0 & 59.8 & 51.6 & 60.0 & 58.4 & 50.5 & 58.3 & 67.8 & 70.3 & 87.0 \\
ResNet-152\cite{Xiao2018Simple} & val17 & 73.9 & 75.9 & 63.7 & 56.1 & 65.5 & 65.1 & 53.5 & 65.4 & 85.4 & 85.5 & 80.3 \\
MDPN-152-A$^*$ & val17 & 71.6 & 76.7 & 68.6 & 60.6 & 64.4 & 62.6 & 54.3 & \textbf{66.0} & 85.6 & 87.2 & 79.1 \\
\hline
MDPN-152-A$^*$ & val18 & 50.9 & 55.5 & 54.0 & 49.0 & 48.7 & 50.5 & 45.1 & 50.6 & 85.7 & 74.0 & 80.3 \\
\hline
\end{tabular}
\end{table*}

\begin{table}[htb]\tiny
\caption{Results on PoseTrack ECCV 2018 Challenge without (Top) and with (Bottom) extra training datasets. * means with tracking. Our methods are \textbf{bold}. }
\label{table_result_challenge}
\centering
\begin{tabular}{c|c|c|c|c|c}
\hline
Methods & Extra? & Wrist mAP & Ankle mAP & Total mAP & Total MOTA \\
\hline
\textbf{MDPN} & No & \textbf{74.5} & \textbf{69.0} & \textbf{76.4} & - \\
\textbf{MDPN$^*$} & No & 69.5 & 66.1 & 72.6 & \textbf{58.5} \\
openSVAI & No & 66.8 & 62.4 & 69.4 & - \\
Loomo & No & 66.4 & 61.8 & 68.5 & 26.9 \\
MIPAL & No & 60.2 & 56.9 & 67.8 & 54.9 \\
\textbf{AlphaPose++} & No & 66.2 & 65.0 & 67.6 & - \\
E2E$^*$ & No & 62.1 & 58.3 & 63.3 & 53.6 \\
openSVAI$^*$ & No & 59.2 & 56.7 & 63.1 & 54.5 \\
\hline
DGDBQ & Yes & \textbf{77.8} & \textbf{75.4} & \textbf{79.0} & - \\
ALG & Yes & 72.6 & 71.1 & 74.9 & 60.8 \\
MSRA & Yes & 73.0 & 69.1 & 74.0 & \textbf{61.4} \\
Miracle & Yes & 68.2 & 66.1 & 70.9 & 57.4 \\
E2E & Yes & 67.0 & 62.5 & 67.8 & - \\
\hline
\end{tabular}
\end{table}

\section{Conclusions}
In conclusion, we investigate the strategies for training on multi-dataset and present   Multi-Domain Pose Network to improve human pose estimation. It surpasses the baselines and achieves state-of-the-art results on PoseTrack benchmarks. Because of the simplicity, we hope proposed methods can help improve the performance for training on multiple datasets.

%
%
%
\bibliographystyle{splncs04}
\bibliography{eccv}
\end{document}